\documentclass[11pt,a4paper]{article}
\usepackage[hyperref]{acl2021}
\usepackage{times}
\usepackage{latexsym}

\usepackage{booktabs}
\usepackage{multirow}
\usepackage{amsmath}
\usepackage{tikz}
\usepackage{tcolorbox}
\usepackage{xcolor}
\usepackage{listings}
\usepackage[scaled=0.8]{beramono}
\usepackage{enumitem}
\aclfinalcopy

\usepackage{microtype}

\title{Is neural semantic parsing good at ellipsis resolution, or isn't it?}

\author{Xiao Zhang \\
  Center for Language and Cognition\\
  University of Groningen \\
  \texttt{xiao.zhang@rug.nl} \\
  \And
  Johan Bos\\
  Center for Language and Cognition\\
  University of Groningen \\
  \texttt{johan.bos@rug.nl}
}

\date{}

\begin{document}
\maketitle
\begin{abstract}
Neural semantic parsers have shown good overall performance for a variety of linguistic phenomena, reaching semantic matching scores of more than 90\%.
But how do such parsers perform on strongly context-sensitive phenomena, where large pieces of semantic information need to be duplicated to form a meaningful semantic representation? A case in point is English verb phrase ellipsis, a construct where entire verb phrases can be abbreviated by a single auxiliary verb. Are the otherwise known as powerful semantic parsers able to deal with ellipsis or aren't they? We constructed a corpus of 120 cases of ellipsis with their fully resolved meaning representation and used this as a challenge set for a large battery of neural semantic parsers. Although these parsers performed very well on the standard test set, they failed in the instances with ellipsis. Data augmentation
helped improve the parsing results. The reason for the difficulty of parsing elided phrases is not that copying semantic material is hard, but that they usually occur in linguistically complicated contexts, causing most of the parsing errors.
\end{abstract}

\section{Introduction}

Semantic parsing is the task of providing a formal meaning representation for an input sentence of a natural language such as English, Dutch, or Italian. Semantic parsing is crucial for applications that require the precise translation of unstructured data (i.e., text and images) into structured data (e.g., databases and robot commands). Currently, the most promising approaches to semantic parsing are based on neural models \cite{bai-etal-2022-graph,wang2023plmp,zhang-etal-2024-gaining,zhang-etal-2025-neural} trained or fine-tuned on large semantically annotated corpora \cite{amr,pmb}, reaching high performance with F scores greater than 90\%. Little is known about the ability of neural semantic parsers to cope with \emph{ellipsis}, a linguistic construction in which elements are omitted and are supplied by the discourse context. In this paper, we will study how neural semantic parsers deal with Verb Phrase Ellipsis (VPE) in English. An example of a VPE is shown in (1) together with its fully expressed surface interpretation in (2).

\smallskip\noindent
(1) Ann likes grapes, and Bea does, too. \\
(2) Ann likes grapes, and Bea does \textbf{like grapes}, too.
\smallskip
\vspace*{-8pt}

\noindent
As this very simple example already demonstrates, ellipsis interpretation is a challenging task, for the only way to recover the elided material is to consider the discourse context. The (computational) linguistics literature abounds with many more complicated examples of VPE, including sloppy-strict interpretation of pronouns appearing in the elided material, cascaded ellipsis, antecedent contained deletion, gapping, and embedded ellipsis \cite{dahl1973so, Williams1977, roberts:modal,dalrymple:ellipsis}. Nevertheless, our aim is not to focus on these linguistically interesting examples carefully crafted by linguists, but rather to investigate how data-driven semantic parsers deal with instances of VPE found in corpora. 

As far as we know, this is the first in-depth study of VPE interpretation in neural semantic parsing. Related, but taking a different perspective, is work by \citet{Hardt:2023}, who found that large language models have difficulty processing ellipsis.

In Section~\ref{sec:background} we give an overview of earlier computational approaches to VPE. In Section~\ref{sec:pmb} we introduce the Parallel Meaning Bank (PMB) and a VPE challenge test set distilled from the PMB. In Section~\ref{sec:exp} we outline our approach to enhance the semantic parsing for VPE, while in Section~\ref{sec:results}  the parsing results are presented, showing that neural approaches face a difficult time in interpreting elliptical constructions, even with substantial fine-tuning, but not for the reasons we initially thought would cause the difficulty.

\section{Background}\label{sec:background}

VPE interpretation has drawn considerable attention in formal linguistics \cite{dahl1973so,sag:phd,klein:vpe,dalrymple:ellipsis}. These early approaches can be summarized as identifying an antecedent verb phrase in the context, providing a logical form while abstracting over the subject, and applying the result to the subject noun phrase of the elided verb phrase.
Computational approaches were introduced later \cite{alshawi:cle,Kehler:1993,Bos1994COLING,crouch:eacl95,hardt:1997}, with the landmark paper by \citet{dalrymple:ellipsis} introducing a set of benchmark VPE examples and a sophisticated algorithm based on higher-order unification to construct fully resolved meaning representations for elliptical phrases. These approaches, although computational of nature, still required external modules to identify the source verb phrase and the parallel elements between source and target phrase. 

Data-driven approaches based on annotated corpora \cite{nielsen:phd,BosSpenader2011,Bos2016Nerbonne} demonstrated the large gap between theoretical ideas and practical implementations \cite{McShane_Babkin_2016,kenyon-dean-etal-2020-deconstructing,Zhang_Zhang_Liu_Di_Liu_2019}, and were considered to be specific tasks rather than an integral part of wide-coverage semantic parsing. In this paper, we take a different computational perspective and depart with an overall well-performing general-purpose semantic parsing and investigate how well it succeeds on ellipsis data.

\section{Data}\label{sec:pmb}

\begin{figure}
    \centering
    \fbox{\includegraphics[width=0.95\linewidth]{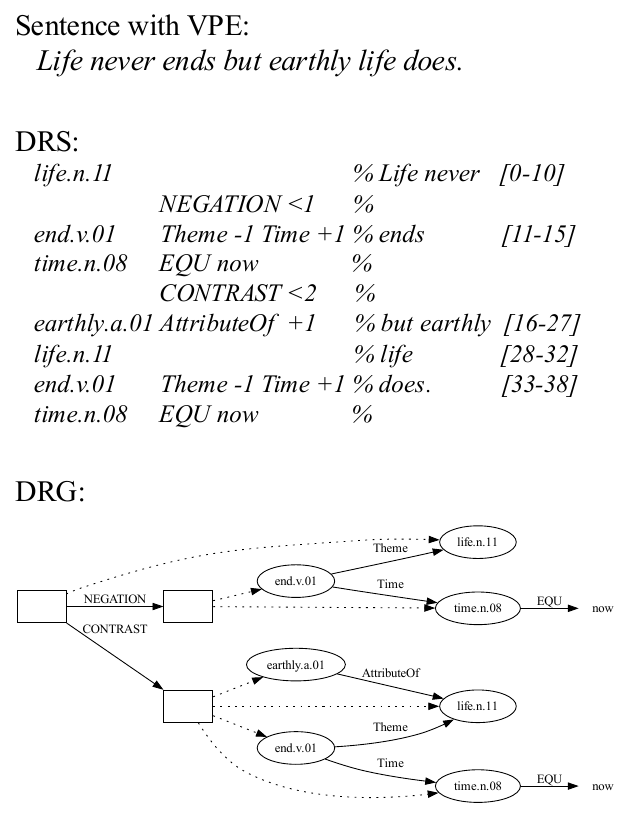}}
    \caption{An example sentence with Verb Phrase Ellipsis and  meaning representation in sequence notation and drawn as a directed acyclic graph.}
    \label{fig:vpe}
\end{figure}

\paragraph{The Parallel Meaning Bank} 
The PMB \citealp{pmb} is a multilingual corpus enriched with semantic annotations, covering a wide range of linguistic phenomena. It contains a substantial set of parallel texts, each paired with a formal meaning representation known as a Discourse Representation Structure (DRS) based on Discourse Representation Theory (DRT, \citealp{kampreyle:drt}). While DRSs are typically presented in a human-readable box format, a clause-based linear representation was introduced by \citet{van-noord-etal-2018-exploring} to enable their use in sequence-based models. More recently, \citet{Bos2023IWCS} proposed Sequence Box Notation (SBN), a simplified, variable-free version of DRS aimed at further facilitating sequence processing. In this paper we use SBN as meaning representation format (see Figure~\ref{fig:vpe}).


\paragraph{Annotated VPE Instances} 

As occurrences of VPE are relatively rare \cite{BosSpenader2011}, it is rather challenging to yield a reasonably sized corpus. A total of 120 cases were identified in the PMB and their corresponding meaning representations manually corrected.
Slightly more than half of the cases (71) contained some kind of negation in the elided construction (e.g., "and neither am I", "Greenland is not", "but she didn't"). Half of the instances are accompanied by the auxiliary verb \emph{to do}, a third by \emph{to be}, and the remaining cases are formed by other auxiliary verbs, the infinitival particle \emph{to} or instances of gapping. An annotated example taken from the corpus is shown in Figure~\ref{fig:vpe}, where the elliptical phrase "life does" is semantically interpreted as "life does end". 

\section{Experimental Setup\label{sec:exp}}

\paragraph{Training Sets}

For training our neural semantic parsers, we consider two settings:  
(1) the \textbf{Standard Training Set}, the default training data provided by PMB version 5.1.0, with all texts that are included in the VPE test set removed; and (2) the \textbf{Augmented Training Set}, an augmented dataset to enhance the model’s ability to handle verb phrase ellipsis. 
We construct the augmented dataset applying the following data augmentation strategies:

\begin{itemize}

    \item We employ GPT-4 to generate 600 sentence pairs, each consisting of a sentence containing VPE and its corresponding resolved version (i.e., the full sentence with the elided verb phrase explicitly restored in the surface text).
    
    \item We use the state-of-the-art DRS parser from~\citet{zhang2024retrieval} to generate DRSs for the resolved sentences. These DRSs are then paired with the original VPE sentences as their target semantic representations.
    
    \item We incorporate the generated VPE data into the standard training set in varying quantities (from 100 up to 600) to examine how the scale of augmentation affects model performance and to identify the point at which performance improvements begin to converge.

\end{itemize}

\paragraph{Test sets}

We evaluate the trained parsers on two test sets: the \textbf{Standard Test Set}, which serves as a general, broad-coverage set for comparison, and \textbf{VPE120}, a targeted test set focusing on VPE, as described in Section~\ref{sec:pmb}.

\paragraph{Evaluation}

We evaluate model performance using two metrics: \textbf{Smatch}\footnote{We adopt the Smatch++ implementation \cite{opitz-2023-smatch}, which uses Integer Linear Programming (ILP) instead of the standard hill-climbing approach.} and \textbf{Ill-Formed Rate (IFR)}. Smatch~\cite{cai-knight-2013-smatch, opitz-2023-smatch} measures the similarity between the predicted and reference semantic graphs by converting each graph into a set of triples and computing the optimal variable mapping via a hill-climbing algorithm. Precision (P), recall (R), and F1 score are calculated as follows:
\begin{equation}
\begin{aligned}
\text{P} &= \frac{m}{p}, \quad \text{R} = \frac{m}{g}, \quad \text{F1} = \frac{2 \cdot \text{P} \cdot \text{R}}{\text{P} + \text{R}},
\end{aligned}
\end{equation}
where \( m \) denotes the number of matching triples, \( p \) is the number of predicted triples, and \( g \) is the number of gold-standard triples.

To assess the structural validity of generated graphs, we additionally report the \textbf{Ill-Formed Rate (IFR)}. A graph is considered ill-formed if it exhibits structural defects such as cyclic dependencies, isolated nodes, or dangling edges referencing non-existent elements. Graphs identified as ill-formed are assigned a Smatch score and F1 score of zero, thereby contributing to a quantitative measure of structural failure.

\paragraph{Models}
We evaluated three encoder–decoder models--mBART \cite{liu-etal-2020-multilingual-denoising}, mT5 \cite{xue-etal-2021-mt5}, and ByT5 \cite{xue-etal-2022-byt5}, as well as four decoder-only models: Qwen2.5-7B \cite{qwen2025qwen25technicalreport}, Ministral-8B, LLaMA3.1-8B \cite{grattafiori2024llama3herdmodels}, and Gemma2-9B \cite{gemmateam2024gemma2improvingopen}.

\section{Results and Analysis}\label{sec:results}


The performance of models on both test sets is presented in Table~\ref{tab:results}. 
Overall, sentences containing VPE instances pose significantly greater challenges for semantic parsing, as evidenced by substantially lower Smatch scores and elevated ill-formed rates (IFR). We analyze these results in detail below.

\begin{table}[htbp]
\centering
\caption{Smatch and IFR performance on the Standard Test Set and VPE120 for models trained with the Standard Training Set, Aug300, and Aug600.}
\label{tab:results}
\small
\setlength{\tabcolsep}{3pt}
\begin{tabular}{llrrrr}
\toprule
\multirow{2}{*}{Model} & \multirow{2}{*}{Train set}
  & \multicolumn{2}{c}{Standard Test}
  & \multicolumn{2}{c}{VPE120} \\
\cmidrule(lr){3-4} \cmidrule(lr){5-6}
                       &           
  & Smatch & IFR   & Smatch & IFR   \\
\midrule
mBart-Large            & Standard & 83.50 & 6.95  & 70.90 & 33.17 \\
                       & Aug300   & 85.40 & 7.00  & 77.90 & 27.33 \\
                       & Aug600   & 85.00 & 6.60  & 78.10 & 24.83 \\
\addlinespace
mT5-Large              & Standard & 82.61 & 11.20 & 70.38 & 29.83 \\
                       & Aug300   & 84.50 & 9.80  & 75.20 & 24.83 \\
                       & Aug600   & 84.00 & 9.20  & 75.50 & 24.00 \\
\addlinespace
ByT5-Large             & Standard & 91.40 & 8.73  & 66.22 & 27.33 \\
                       & Aug300   & 92.50 & 7.50  & 73.00 & 22.33 \\
                       & Aug600   & 92.90 & 7.00  & 72.50 & 22.33 \\
\addlinespace
Qwen2.5-7B             & Standard & 94.19 & 5.34  & 77.09 & 17.33 \\
                       & Aug300   & 94.35 & 5.17  & 85.31 &  9.83 \\
                       & Aug600   & 95.50 & 5.09  & 84.64 & 12.33 \\
\addlinespace
Ministral-8B           & Standard & 95.45 & 4.67  & 82.77 & 13.17 \\
                       & Aug300   & 95.50 & 4.25  & 89.00 &  6.50 \\
                       & Aug600   & 95.42 & 4.59  & 90.61 &  6.50 \\
\addlinespace
LLaMA3.1-8B            & Standard & 95.56 & 4.51  & 83.11 & 12.33 \\
                       & Aug300   & 95.32 & 5.18  & 88.89 & 12.33 \\
                       & Aug600   & 95.44 & 4.76  & 89.21 &  8.17 \\
\addlinespace
Gemma2-9B              & Standard & 96.31 & 4.59  & 78.09 & 17.33 \\
                       & Aug300   & 96.46 & 4.42  & 88.52 &  7.33 \\
                       & Aug600   & 96.59 & 4.09  & 89.20 &  8.17 \\
\bottomrule
\end{tabular}
\end{table}

\paragraph{Performances on Standard Test}

Decoder-only architectures consistently outperform encoder--decoder models on the Standard Test set. Gemma2-9B achieves the highest performance with a Smatch score of 96.59 following augmentation (compared to 96.31 on the standard training set). Other decoder-only models demonstrate similarly strong performance: LLaMA3.1-8B (95.44), Ministral-8B (95.50), and Qwen2.5-7B (95.50) all maintain scores above 95. In contrast, encoder--decoder architectures (mBART-Large, mT5-Large, and ByT5-Large) achieve lower performance, with \emph{standard} scores ranging from 82.61 to 91.40. This performance disparity likely stems from both architectural differences and parameter scale advantages, where larger decoder-only models may benefit from more stable fine-tuning dynamics and in-context learning capabilities.

\paragraph{Performances on VPE120}

All models exhibit substantially degraded performance on VPE120 relative to the Standard Test, confirming the inherent difficulty of parsing elliptical constructions semantically. When trained solely on the standard dataset, models achieve VPE120 Smatch scores between 66.22 and 83.11, accompanied by markedly increased IFR (e.g., 33.17\% for mBART-Large and 29.83\% for mT5-Large), indicating frequent generation of malformed outputs.

VPE-specific data augmentation yields substantial improvements across all architectures. Ministral-8B achieves the highest score of 90.61 with Aug600, closely followed by LLaMA3.1-8B (89.21) and Gemma2-9B (89.20). These top-performing models also demonstrate the most significant IFR reductions (e.g., Ministral-8B: 13.17\% $\rightarrow$ 6.50\%). Encoder--decoder models also benefit from augmentation: mBART-Large improves from 70.90 (standard) to 78.10 (Aug600), while mT5-Large advances from 70.38 to 75.50. Notably, ByT5-Large shows improvement from 66.22 to 73.00 with Aug300.

\begin{figure}[htbp]
    \centering
    \includegraphics[width=\linewidth]{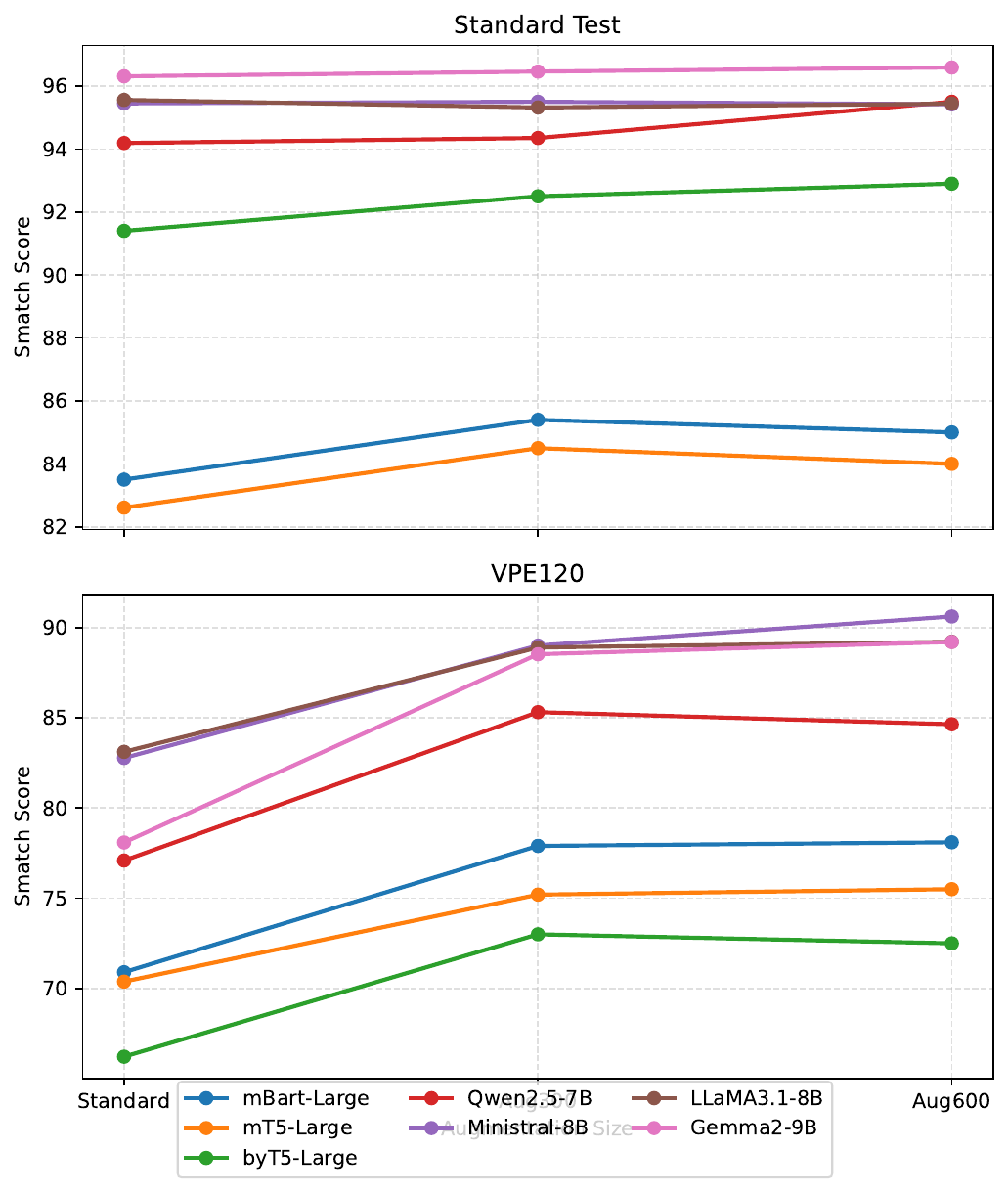}
    \caption{Model performance on VPE120 with increasing augmentation sizes (100 to 600).}
    \label{fig:results}
\end{figure}

These findings demonstrate that VPE-specific data augmentation effectively narrows the performance gap between the Standard and VPE test sets, particularly for larger decoder-only models. The convergence of performance scores beyond Aug300 (see Figure~\ref{fig:results}) suggests diminishing gains from additional augmentation data, indicating that current models may be approaching their capacity limits for ellipsis resolution. This shows the need for more advanced architectures or specialized training strategies to further improve performance on complex elliptical phenomena.

\paragraph{Qualitative Analysis}

The previous section showed that sentences with ellipsis are a lot harder to parse for the neural semantic models. But why is this the case? Is this because they are a bad at copying semantic information, or is it something else? In order to answer we examined the output of the best performing model and manually inspected the results. 

Surprisingly, what we thought would be hard for the models, copying semantic material from the source to the target, was not hard at all. Only in three of the 120 cases did this not happen. Actually, what contributed to the low score was 
the wrong choice of discourse relation (22\% of overall errors), 
the wrong attachment of a discourse relation (20\%), 
incorrect scope order between tense and negation (16\%), 
incorrect choice of word sense (16\%), 
incorrect choice of thematic role (10\%),
incorrect choice of concept (10\%), and incorrectly resolved anaphora (4\%).

One reason why selecting the correct VP antecedent might have to do with the amount of ambiguity, or lack thereof. For instance, in the VPE example in Figure~\ref{fig:vpe} there is only one potential verb phrase that could serve as antecedent for the elliptical phrase. Closer inspection of the dataset reveals that most (81\%) of the texts with VPE are relatively short and provide only one verb phrase that could act as antecedent; only 23 examples provide two or more potential verb phrase antecedents, as in (3). 

\smallskip\noindent
(3) Ann hoped to succeed, but she didn't. 
\smallskip

\noindent 
Here there are two verb phrases in the context: \emph{hope to succeed} and \emph{succeed}. For most of these cases picking the most recent verb phrase usually yields the correct interpretation.

\section{Conclusion}\label{sec:conclusion}

Although open-domain semantic parsing achieves good overall performance on the standard test sets, its shortcomings arise at the surface when looking at more complex linguistic phenomena. We demonstrated this by looking specifically at how neural parsing models deal with cases of English VP Ellipsis. Although we observed a drop in performance, the reason for the drop was not the context-sensitive nature of ellipsis, but rather the fact that elliptical phenomena are often surrounded by complex phenomena such as tense, negation, and discourse structure, causing parsing errors. So, is neural semantic parsing good at ellipsis resolution? Yes, it is! 

\bigskip

\subsection*{Acknowledgements}

We would like to thank the three anonymous reviewers for their comments.  
Reviewer 1 pointed out that the developed datasets will be valuable for the community, and indeed the VPE dataset will be made public via the Parallel Meaning Bank data releases. 
Reviewer 2 wondered how many different VP targets there are for each case of ellipsis, as the complexity for ellipsis resolution "depends greatly on the number of available VP targets for each VPE". We added a discussion in this topic in Section~\ref{sec:results}.
Reviewer 3 noted that the question in the paper's title is not actually answered. This was a correct observation. But now we did explicitly in Section~\ref{sec:conclusion}.
Special thanks go to Juri Opitz, who pointed out that using a hill-climber for evaluation (as we did in the submitted version of this paper) is not optimal. So we recalculated our scores using ILP instead. 

\bibliographystyle{acl_natbib}
\bibliography{anthology,acl2021,comsem}

\appendix

\end{document}